# Comparison of Deterministic and Probabilistic Machine Learning Algorithms for Precise Dimensional Control and Uncertainty Quantification in Additive Manufacturing

Dipayan Sanpui[1,2], Anirban Chandra[1,2,*], Henry Chan[1,2], Sukriti Manna[1,2] and Subramanian K.R.S. Sankaranarayanan[1,2]

[1] Center for Nanoscale Materials, Argonne National Laboratory, Lemont, Illinois 60439, United States.
[2] Department of Mechanical and Industrial Engineering, University of Illinois, Chicago, Illinois 60607, United States.
*Currently at Shell International Exploration and Production Inc., Boston, Massachusetts, 02210, United States

## Abstract

We present an accurate estimation of the dimensions of additively manufactured components by adopting a probabilistic perspective. Our study utilizes a previously gathered experimental dataset, encompassing five crucial design features for 405 parts produced in nine production runs. These runs involved two machines, three polymer materials, and two-part design configurations. To illustrate design information and manufacturing conditions, we employ data models that integrate both continuous and categorical factors. For predicting Difference from Target (DFT) values, we employ two machine learning approaches: deterministic models that offer point estimates and probabilistic models generating probability distributions. The deterministic models, trained on 80% of the data using Support Vector Regression (SVR), exhibit high accuracy, with the SVR model demonstrating precision close to the process repeatability. To address systematic deviations, we introduce probabilistic machine learning methodologies, namely Gaussian Process Regression (GPR) and Probabilistic Bayesian Neural Networks (BNNs). While the GPR model shows high accuracy in predicting feature geometry dimensions, the BNNs aim to capture both aleatoric and epistemic uncertainties. We explore two approaches within BNNs, with the second approach providing a more comprehensive understanding of uncertainties but showing lower accuracy in predicting feature geometry dimensions. Emphasizing the importance of quantifying epistemic uncertainty in machine learning models, we highlight its role in robust decision-making, risk assessment, and model improvement. We discuss the trade-offs between BNNs and GPR, considering factors such as interpretability and computational efficiency. The choice between these models depends on analytical needs, striking a balance between predictive accuracy, interpretability, and computational constraints. In summary, our analysis of an additive manufacturing dataset through the lens of a Probabilistic Bayesian Neural Network (BNN) and the simultaneous quantification of both epistemic and aleatoric uncertainties provides a robust foundation for advancing manufacturing design.

# List of Abbreviations

| Abbreviations | Meaning |
| --- | --- |
| DLS | Digital Light Synthesis |
| DFT | Difference From Target |
| BNN | Bayesian Neural Network |
| AM | Additive Manufacturing |
| SPC | Statistical Process Control |
| DOE | Design of Experiments |
| MCMC | Markov Chain Monte Carlo |
| VI | Variational Inference |
| SVR | Support Vector Regression |
| XGBoost | Xtreme Gradient Boosting |
| GPR | Gaussian Process Regression |
| RMSE | Root Mean Squared Error |
| MSE | Mean Squared Error |
| RF | Random Forest |
| LGBM | Light Gradient Boosting |
| MLP | Multi-layer Perceptron |
| ML | Machine Learning |
| SHAP | Shapley Additive Explanations |
| NN | Neural Network |
| UMA | Urethane Methacrylate |
| EPX | Additive Epoxy |
| RPU | Rigid Polyurethane |
| GP | Gaussian Process |
| MC Dropout | Monte Carlo Dropout |
| UQ | Uncertainty Quantification |

# Introduction

The decision-making process of acceptance or rejection of produced part is critical in a production pipeline and mainly depends on the dimensional accuracy. The additive manufacturing techniques, useful in fabrication of intricate geometries; is highly flexible and undergoes frequent parametric variations. Oftentimes, traditional methods [1-5] for measurement of the produced parts are time-consuming and incur higher monetary investments. While the conventional methods for part quality assessment are sometimes challenging, recent advancements in measurement techniques that involve automated measurements and real-time analysis of data offers significant benefits in the measurement process; often termed as smart metrology-based approach [6].

An existing challenge in the smart metrology-based approach is handling a high-dimensional dataset having uncorrelated processing conditions. In the recent past, data-driven methods have been successful in unleashing the interrelations between manufacturing conditions and dimensional accuracy [7]. Amongst the data-driven methods, deterministic regression methods [8-11], are popular in dimensional accuracy prediction since the implementation is quite straightforward and computationally efficient. However, possess certain limitations while addressing the uncertainty inherent within the measurements. The lack of reliable uncertainty estimates hinders the direct use of the deterministic machine learning algorithms, for manufacturing and materials science-based applications.

It is imperative to estimate uncertainties to provide scientific and engineering decision-makers with predictive information as well as quantitative information regarding how accurate the predictions are. GUM (Guide to the expression of Uncertainty in Measurement) [12,13] is an internationally accepted master document that provides guidelines to assess the uncertainties associated with various sources of error in measurements. The steps for the uncertainty budget include the sources of uncertainties, classification, determination of standard uncertainty and combining them. Another two distinct approaches for uncertainty evaluation are the propagation of distributions through Monte Carlo approaches [12] and Bayesian uncertainty evaluation [13]. Propagation of distributions through Monte Carlo evaluation is a method for determination of uncertainty when the inputs are uncertain and randomly sampled for propagation of uncertainty within the model itself. Instead of relying solely on the analytical methods, the Monte Carlo approaches use random sampling for propagation of uncertainty, thus allowing for a more detailed and accurate characterization of output distribution. While the Bayesian uncertainty evaluation relies on the prior beliefs or knowledge, in most cases considered as standard normal distributions followed by updating the knowledge with new data and provide a full probability distribution for the quantity of interest.

The outcomes of scientific applications including in advanced manufacturing are uncertain, both on an aleatoric and epistemic level [17-19]. Aleatoric uncertainty refers to the uncertainty present in the dataset itself and arises due to stochastic experimental design and noise present in the experimental output. Epistemic uncertainty, also known as subjective uncertainty or knowledge uncertainty, refers to the uncertainty arising from limitations in our understanding, knowledge, or information about a system or phenomenon [20]. This type of uncertainty reflects the lack of full knowledge about the true nature of the system, including its behavior, parameters, or underlying mechanisms. Epistemic uncertainty can manifest in various forms, such as parameter uncertainty that arise due to the model parameters, model uncertainty that occurs due to the model architecture and the input uncertainty, that occurs due to the choice of input features and boundary conditions. Besides aleatoric uncertainty, which is inherent in the data itself, machine learning applications, in particular, suffer from epistemic uncertainties that arise from a lack of knowledge or data from experiments and model hyperparameters.

Based on the theoretical approaches [12,13], relatively a newer perspective towards evaluation of uncertainty is by utilization of the ML algorithms. Epistemic uncertainty can be reduced by providing the ML model with more data [21]. In contrast with the primary ML algorithms, that depends on a lot of information, engineering applications frequently manage limited quantities of complicated information, bringing about enormous epistemic uncertainties. Therefore, the decision maker must characterize both the uncertainties, aleatoric and epistemic while making predictions. In this context, probabilistic ML models serve the purpose of the prediction of both uncertainties [22]. A widely known ML algorithm that incorporates uncertain predictions is Gaussian processes (GPs), inherently derived from the Bayesian learning framework [23]. In recent years, the integration of neural networks with the Bayesian learning framework has become popular among the machine learning research community [24]. Previous literature suggests that there are different ways to incorporate Bayesian inference within neural networks [25]. Bayesian Neural Network (BNN) makes use of Markov Chain Monte Carlo (MCMC) for approximation of the prediction as a probability density function (posterior pdf). Alternatively, the variational inference (VI) approximates the posterior pdfs via distance metrics which can be computed analytically and easy to sample [23]. Another probabilistic approach is MC Dropout [31], which uses neural networks to estimate the uncertainty involved within the model training and alleviate the issue of overfitting. This probabilistic model mimics the Bayesian-like behavior by dropping out neurons in the network randomly. This algorithm is computationally less costly, but several researchers reported the underperformance of this algorithm [18]. The MCMC method usually does not scale well with the model weights and parameters, and thus the convergence is intractable considering the presence of complex integrals for approximation of posterior distributions [31]. VI frameworks integrated with neural networks assume a form of the posterior distribution after each epoch and are thus less accurate than MCMC in their estimation of the posterior distribution, but they are typically faster via optimization; thus, easily leveraged with neural network training, and convergence is fast. Currently, uncertainty quantification for additive manufacturing dimensional predictions lacks extensive exploration using probabilistic regression algorithms [32]. Incorporating Probabilistic BNN-based regression into uncertainty quantification efforts can address the gaps by providing a powerful tool for modeling complex relationships, handling uncertainty, and offering real-time adaptability and interpretability in the context of additive manufacturing dimensional predictions.

In this work, we compare algorithms for probabilistic training of neural networks based on variational Bayesian inference, while using the experimental additive manufacturing dataset [16] for fabrication and measurement of produced parts using Digital Light Synthesis (DLS). We first select diverse non-linear deterministic models for the prediction of dimensional accuracy. While separation of data and model uncertainties are cumbersome using deterministic ML algorithms, probabilistic ML algorithms excel in quantification of both aleatoric and epistemic uncertainty. The remainder of the paper is structured as follows. In Section 2, we present the details of the experimental dataset that we use for the study and describe our implementation techniques for the ML algorithms, followed by Results and Discussions in Section 3. Section 4 concludes with a summary of our observations, concluding remarks, and future scopes.

## 2. Methods

### 2.1 Description of the Dataset

We utilize an additive manufacturing dataset prepared by McGregor et al. [16], that consists of information for the fabrication and measurement of produced parts using the Digital Light Synthesis (DLS) [40] method. The DLS method is a vat-polymerization technique that utilize digital light projection and oxygen permeable optics for continuous fabrication of highly intricate and detailed parts from photosensitive resin. For collection of the dataset and exploration of different manufacturing parameters, McGregor et. al. [16] fabricated a total of 405 parts. Two Carbon M2 printers were utilized to fabricate the parts, each having unique hardware sets (machine ids). Three unique part designs, namely clips, plugs and brackets were manufactured. The build area was divided into three different segments; one-third for each segment for each part design, made with same material. Two organizational layouts were adopted for the experiments, namely, A and B. Parts were arranged in one of the two organizational layouts, in such a manner that the location of each design cluster was different. For example, the layout A was organized with 15 clips on the left side, 15 plugs in the middle, and 15 brackets on the right side. Similarly, the layout B was organized with 15 brackets on the left side, 15 clips in the middle and 15 plugs on the right side. The part designs, i.e.- clips, plugs and brackets were fabricated in a batch process We present a detailed schematic of the experimental design in Fig 1. Therefore, in each batch, 45 parts were manufactured, and 9 experiments were performed that resulted in a total of 405 parts. Five measurements were taken for each manufactured part, which resulted in a total of 2025 measurements. Three different polymer materials were used for fabrication of the produced parts, namely urethane methacrylate (UMA); rigid polyurethane (RPU) and additive epoxy (EPX). In this work, we predict the dimensional accuracy of manufactured parts, the key metric for the prediction is Difference from Target (DFT), which serves as the target variable from the dataset and is the measure of dimensional variations of the produced parts from a reference CAD geometry. Initially during experiments [16, refer supplementary information], duplicate measurements were taken from repeated scans and one average measurement was reported for each of the 7290 measured features. Depending on the design and measured features per part; with subsequent down sampling of features reduced the initial dataset from 7290 measurements to 2025 measurements. In addition to the measured dimensions of five features on each part, the data set also provides with details on the manufacturing parameters and descriptors associated with each feature.

Corresponding to each measured part dimension, there are 13 input features, mixture of continuous and categorical variables. Amongst those 13 input features, eight of them are manufacturing parameters and rest of the inputs are the geometrical descriptors of the measured part. The manufacturing parameters are the hardware settings, material, thermal cure, cartesian and radial co-ordinate locations within the build. Rest of the input features were the measured feature descriptors, i.e.- feature included nominal dimension, feature category, part design, unique feature ID, and feature class. In reference [27], McGregor et al. performed an analysis to study the relative importance of input features and their contributions towards feature DFT predictions. Amongst the total of 13 input features, they observed 8 input features contribute significantly towards the output prediction. Therefore, we choose 8 input features, where six were the manufacturing parameters and rest of the input features were the measured feature descriptors, i.e.- feature class and feature category. We convert the categorical variables into continuous input features using one-hot encoding. The feature category is the topological descriptor for a feature category class, i.e.- inner or outer length of the diameter. The feature class amongst the measured feature descriptor was either thickness, length, diameter, or height. In Fig. 1 and 2, we present an overview of the experimental design and a detailed schematic of the input features and output/target variable along with the part designs respectively.

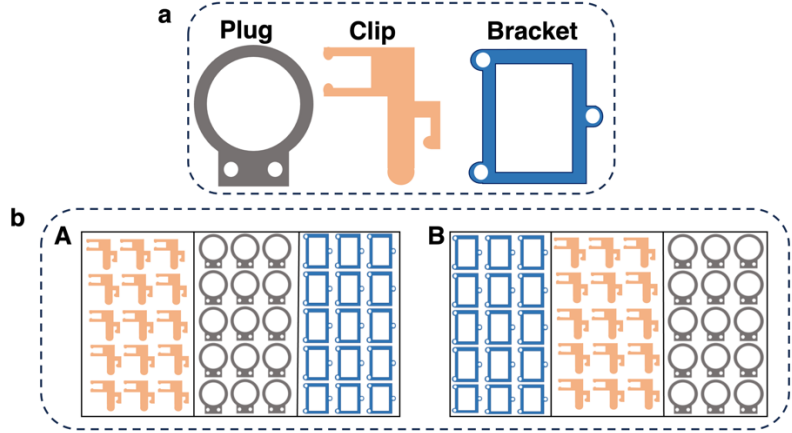

**Fig 1: Representation of the experimental design. a** Three different parts were fabricated namely, plug, clips and bracket, using three different materials, namely urethane methacrylate (UMA); rigid polyurethane (RPU) and additive epoxy (EPX). **b** The build area was divided into three segments, one-third for each part design. Two organizational layouts were adopted namely A and B. In each experiment, the layout A was organized with 15 clips on the left side, 15 plugs on the middle and 15 brackets on the right side. Similarly, the layout B consisted of 15 brackets on the left, 15 clips in the middle and 15 clips on the right side of the build area. Each build area consisted of 45 parts, a total of 9 experiments were performed which produced a total of 405 parts with five measurements for each part, that resulted in a total of 2025 measurements within the dataset.

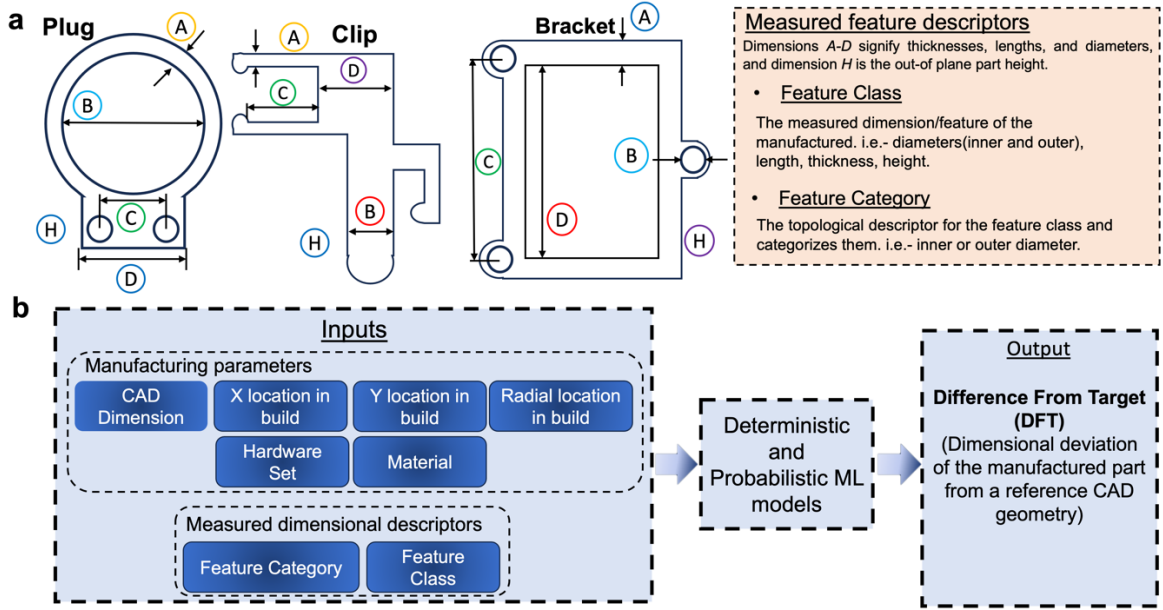

**Fig 2: A detailed representation of the input and output features. a)** Computer Aided Draft (CAD) designs of the additively manufactured parts using the DLS method. Three different part designs with the five critical dimensions/features A-D are shown in the figure. The measured features A-D signify the thickness, length, diameter, and out-of-plane height of the manufactured part. **b)** We use the existing experimental dataset prepared by McGregor et al. [27] and use it for training of the ML models. The inputs for training the ML models consists of a mixture of numerical and categorical variables. For a clear representation we demarcate between the manufacturing parameters and the measured dimensional/feature descriptors describing the topology of the parts. As the output from the ML models, we predict the Difference from Target (DFT), signifies the dimensional deviation from the refence CAD geometry.

Within the realm of advanced manufacturing methods, the development of a machine learning framework to predict the dimensions of additively manufactured built parts involves a supervised regression or classification task [9,10,11]. When provided with a set of input-output data, the objective is to develop a regressor function $f(\cdot)$ that can accurately predict the output $y = f(x)$ for a new input $x$. The input feature vector, such as structured input data, or it may represent an image from a microscope of different phases in a microstructure and corresponding microstructure features, for example, image input data/microstructure feature distributions [14,15]. The output $y$ represents the target feature of interest, the DFT values of the built parts. Our work uses both deterministic and probabilistic ML models for the prediction of the dimensional accuracy of AM parts. We use the dimensions of the feature descriptors along cartesian and radial coordinates as continuous inputs to the model. Manufacturing parameters and measured feature descriptors were the categorical inputs to the models for the prediction of dimensional accuracy.

An important factor to consider is the random variance present within the underlying data that limits the ability of the model to make accurate predictions. For the experimental dataset we utilized, different polymeric materials used for fabrication of the parts possess repeatability. The weighted average of the repeatability of materials utilized for fabrication of the parts is $\pm 47\ \mu m$. Moreover, the measurement uncertainty during curation of the data is $\pm 15\ \mu m$. Besides, the combined estimate of production repeatability and measurement uncertainty is $\pm 50\ \mu m$ (root sum of squares). This estimate indicates that an ideal regression method while evaluated with the test data might achieve a minimum of $50\ \mu m$ root mean squared (RMSE) value for DFT predictions.

Furthermore, for the target feature data distribution, the standard deviation is $180\ \mu m$ and serves as to provide as a baseline for evaluation of the ML regression methods utilized. Smaller prediction accuracy than this baseline value shows the utility of incorporated ML methods. However, a critical fact is that the target feature; dimensional deviations of the measured feature from a reference geometry/Difference from Target (DFT) is analogous to the prediction of random variance inherent in the data; which is a combination of both production repeatability and measurement uncertainty. It is noteworthy, that while we utilize deterministic regression methods for DFT predictions, we actually predict the random variance within the data, which can be attributed to aleatoric uncertainty.

During experiments, in real-time manufacturing conditions, there are some factors that affects the process itself. For instance, an unrecognized factor like a calibration error in a machine causes consistent deviations (bias) in production outcomes. Similarly, this scenario can be compared with a probabilistic regression method that learns distributions over weights and quantifies uncertainty based on the model hyperparameters and amount of data provided. Both of the scenarios can be attributed to systematic variance/biases. We explicitly try to evaluate the aleatoric (random variance) and epistemic (systematic variance) uncertainties by implementing Bayesian Neural Networks.

Fig.3 shows a consolidated representation of framework, which we implement for the training and evaluation of multiple ML algorithms. An ML algorithm is the underlying computational method used to identify interdependencies between data and predictions; for example, Random Forest (RF) is an ML algorithm. It is noteworthy that, ML model is an instantiation of an algorithm with a particular set of hyperparameters and subject to the distinctive data that is sampled for training.

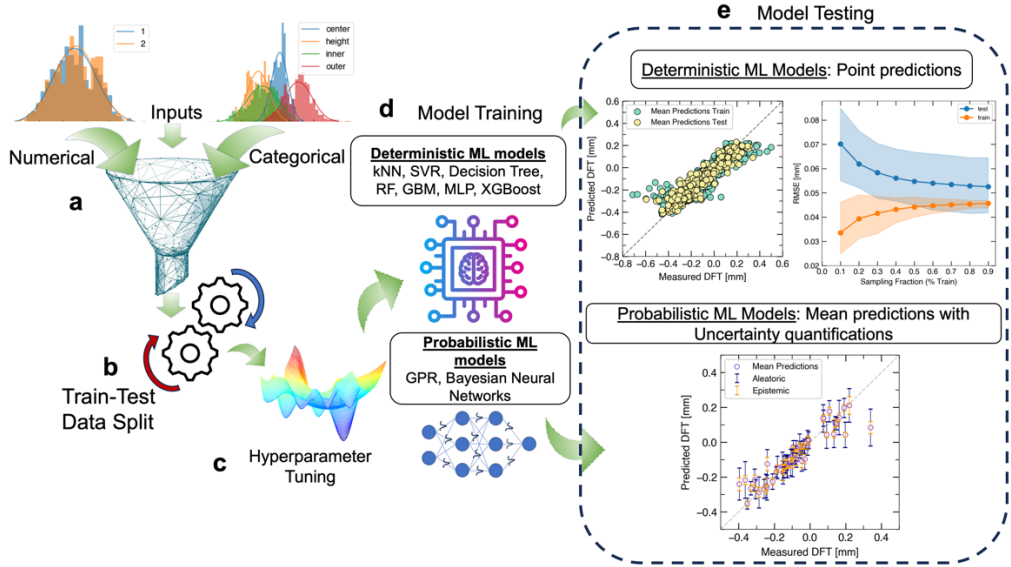

**Fig 3: A consolidated representation of the ML methods we use for our work**. **a, b, c** We use the existing experimental dataset prepared by McGregor et al. [27] for the DLS method and utilize it for training of the ML models. The inputs for training the ML models consists of both numerical and categorical variables followed by hyperparameter tuning of parameters. **d** We use both deterministic and probabilistic ML models for training purpose. **e** Furthermore, we test the trained deterministic models, amongst which SVR and XGBoost shows nearly identical performance and gives a point estimate as prediction. While testing the probabilistic ML models we use two different ML models that quantify the measurement uncertainty, while the GPR model can only characterize the aleatoric uncertainty, BNN's are able to characterize both the aleatoric and epistemic uncertainty.

**2.2 Deterministic ML algorithms**

A similar approach was followed for the implementation of deterministic models as described in Ref [16]. We utilize Dual Monte Carlo subsampling method comprises of two nested loops and subsequent steps to process data, including subsampling/data splitting, normalization and tuning of hyperparameters. In the outer loop, the data is randomly split into training, test and unused/holdout sets, ensuring that the training data is not used for testing purposes and the withheld unused split for test data efficiency evaluation. The inner loop utilizes nested k-fold cross-validation of data over choosing the regular k-fold cross-validation. The key advantage of the nested k-fold cross-validation over the regular method is reduced overfitting, unbiased performance estimation, better hyperparameter tuning, and reliable model comparison. Thereafter, we use the hyperparameter-tuned model to predict the geometry of unmeasured features using the test data. The inner loop iterates multiple times (50 iterations), training and evaluating a new model at each stage. After performing multiple iterations, we average over all the expectations of model accuracies at each iteration and estimate the overall performance (evaluation based on the withheld test set and reported as RMSE) of the model for a distinctive set of inputs. Reshuffling of the data was performed in the outer loop (3 iterations) and the process repeats itself through the inner loop. In section 1 of SI, we elaboratively discuss the dual Monte Carlo subsampling method, along with the pictorial representation of the method in Fig. S1.

Initially, we evaluate seven deterministic regression algorithms using Scikit-Learn [28], a Python package for machine learning, for the prediction of the feature DFT of additively manufactured parts. The different deterministic ML algorithms that we used are k-nearest neighbors, Support Vector Regression (SVR), decision trees, and generalized decision tree-based algorithms such as Random Forest, Gradient Boosting, Extreme Gradient Boosting (XGBoost) [29,30] and the multi-layer perceptron (MLP) regressor [31]. In section 3 of SI, we discuss the different deterministic ML algorithms, their working procedure and the hyperparameters we use to train the models.

We report the best results from the SVR and XGBoost algorithms, as they provide us with nearly identical performance; discussed in more detail in the results section. SVR is usually known for its widespread use, however, XGBoost, a scalable machine learning system for tree boosting, is relatively new in the machine learning community gives state-of-art results and used for different applications in the recent past [29,30]. The most important factor behind the near-identical performance of XGBoost when compared with SVR is its scalability in all scenarios. The scalability of XGBoost is due to several important systems and algorithmic optimizations [29].

## 2.3 Probabilistic ML Algorithms

We list and describe below the probabilistic ML algorithms used in this study.

### 2.3.1 Gaussian Process Regression

Gaussian Process Regression (GPR), as implemented in the Scikit-Learn [28] library, is a non-parametric machine-learning technique used for regression tasks. It leverages Gaussian processes, which are collections of random variables, to model the relationship between input features and target variables. GPR not only provides point predictions but also quantifies uncertainty in predictions, making it particularly useful in scenarios where understanding and managing prediction uncertainty is crucial. We use a combination of the kernel functions – Matérn and White kernels for our work. For the Matérn kernel, we use the length scale (equals 1) and the smoothness parameter (equals 1.5) as the hyper-parameters. Also, for the White kernel that adds a noise component to the GPR model, we use noise level (equals 1) and the number of optimizer restarts (equals 50) as the hyper-parameters. While the implementation of GPR is straightforward, we get an overall uncertainty estimate, thus separation of uncertainties is not possible. Additionally, scalability is a concern for a vast amount of data, as the computational requirements for training and prediction can become prohibitive. Therefore, as an alternative, for trustworthy uncertainty quantification and scalability, next we use Bayesian learning methods.

### 2.3.2 Bayesian Neural Networks

Different from a standard neural network, Bayesian neural network incorporates additional probabilistic information, and the network weights are represented as distributions. In a traditional neural network (Fig 4(a)), the weights are optimized through the backpropagation mechanism, yields a point estimate provided with a labeled data. Our study centers on a Bayesian deep learning approach (Fig. 4(b)) that implements stochastic variational inference derived neural networks with the integration of Bayesian probability theory [21].

Implementation of Bayesian inference calculates the conditional probability of approximating a posterior distribution given the training data, the approximated posterior distribution calculated is outlined in equation 1[25]:

$$P(\omega|D_{tr}) = \frac{P(\omega|D_{tr})P(\omega)}{P(D_{tr})} = \frac{P(\omega|D_{tr})P(\omega)}{\int P(\omega|D_{tr})P(\omega)d\omega} \quad (1)$$

Where, $\omega$ represent the weights, $D_{tr}$ refers to the training datapoints. $P(\omega)$ signify the prior distribution; $P(\omega|D_{tr})$ represent the likelihood estimation and $P(D_{tr})$ is the evidence.
Given a set of inputs, the output target variable can be probabilistic predicted, presented in equation (2) [25]:

$$P(\bar{y}|\bar{x}, D_{tr}) = \int P(\bar{x}|\bar{y}, \omega) P(\omega|D_{tr}) d\omega \quad (2)$$

$P(\bar{x}|\bar{y}, \omega)$ is the predictive distribution/approximated posterior after each forward pass. $\bar{x}$ is the set of inputs and $\bar{y}$ is the corresponding output distribution.

In Bayesian neural networks, the uncertainty that arise from the model parameters is termed as the epistemic uncertainty, which is quantified by learning the probability distribution over model parameters, i.e.-weights. Initially, a prior distribution is considered before observing the training data. The posterior distribution is then approximated using the Bayesian inference outlined in equation 1. However, derivation of the posterior term directly from the is cumbersome due the presence of intractable multi-dimensional integrals presents in denominator term in equation 1. To address this, Markov Chain Monte Carlo [31] and Variational inference [23] are developed. Considering the high-dimensional non-linear input feature space, we implement variational inference for approximation of the posterior distributions. We utilize TensorFlow-Probability [34] module from TensorFlow [35] for probabilistic reasoning and statistical analysis, to implement the probabilistic Bayesian neural networks.

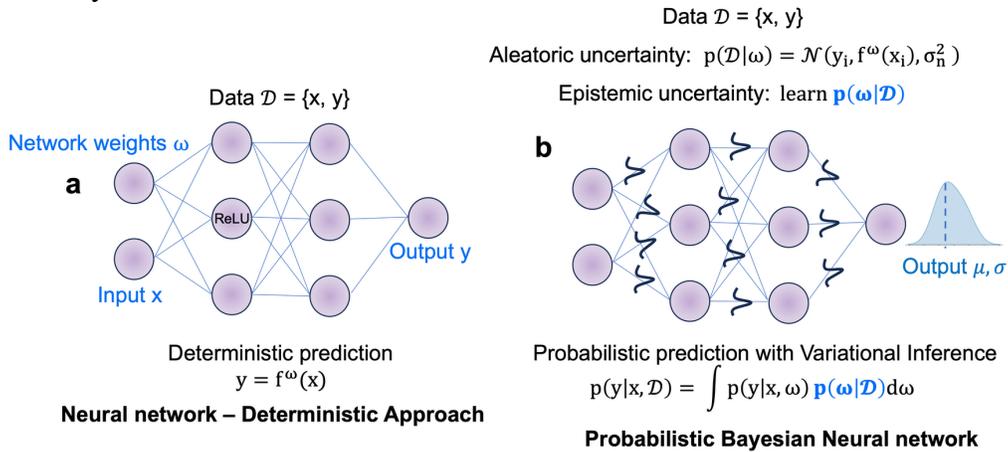

**Fig 4**: **Representation of neural network architectures – deterministic and probabilistic approaches. a** The traditional neural networks (left), as the regression models consisting of layers of elementary non-linear functions, specified using the network weights and biases that minimize the absolute error between the prediction and the true value. Traditional deterministic neural networks do not account for uncertainties in model predictions. **b** Bayesian Neural Networks learns distributions over weights and gives the output as a distribution with a mean and variance. A probabilistic BNN works as an ensemble while we use variational inference. BNN's (right) are designed to account

for both the epistemic and aleatoric uncertainties. We provide the details of the terminologies mentioned in the equations in section 4 of SI.

*2.3.2.1 Bayesian neural network with trainable mean and variance – yields a single output distribution*

As an initial step, we assume a prior distribution with zero mean and unit variance (a one-dimensional tensor) using the 'IndependentNormal' (to treat independent distributions as a single distribution with multiple independent dimensions) layer. Next, we use fully connected layers followed by batch normalization for computation of parameters. We use a total of three hidden layers, while the first hidden layer consisted of 24 neurons, the second one included 16 neurons and the third one included 8 neurons, with Rectified linear unit (ReLU) activation function with each hidden layer. We approximate a posterior distribution using the 'MultivariateNormalTril' layer that define a probabilistic output layer where the output is drawn from a multivariate normal distribution parameterized by the predictions from the fully connected 'Dense' layers with the utilization of prior distribution. Here we define a one-dimensional 'MultivariateNormalTril' distribution that defines 2 parameters (1 for the mean of the distribution and 1 for the variance). We add a Kullback-Leibler (KL) divergence regularization term that penalize the divergence between the learned distribution and the prior distribution, encouraging the model instantiation to maintain proximity between the prior and posterior. The regularization parameter is scaled by the length of training dataset. We present the abovementioned neural network architecture in Fig. 5.

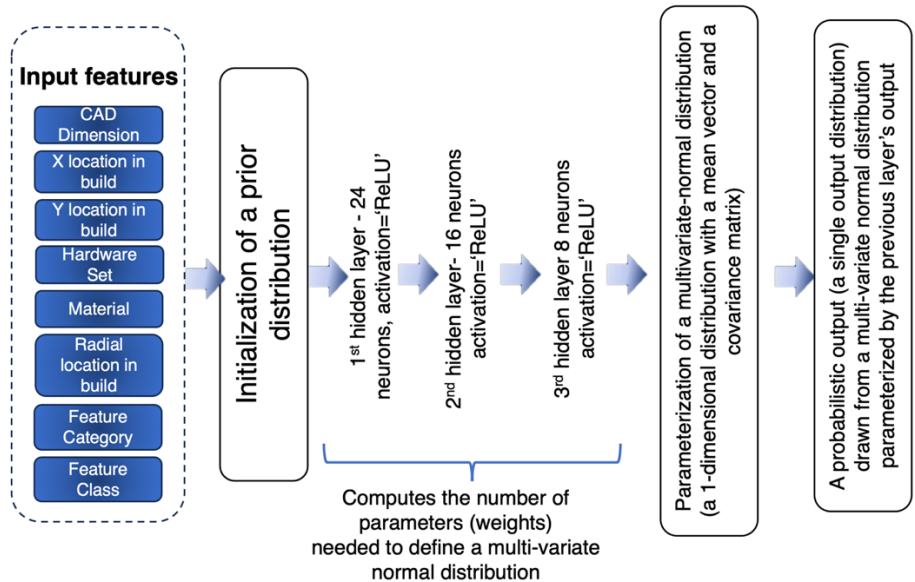

**Fig 5**: **Representation Bayesian neural network with trainable mean and variance – yields a single output distribution**. As an initial step, a prior distribution (having zero mean and unit variance) was initialized followed by computations of parameters (weights and bias) in the intermediate layers. The outputs from the dense layers were used to parameterize a distribution object using the IndependentNormal layer. We choose the event size as one to get a single probabilistic output distribution.

## 2.3.2.2 Bayesian neural network working as ensemble of networks

We start with assuming an initial prior distribution for the model weights expressed by a set of multivariate normal gaussian distributions (having zero means and unit variances), a single multivariate distribution rather than a batch of independent distributions. After each forward pass the posterior distribution is approximated using the "MultivariateNormalTril" layer that make use of the multivariate Gaussian distribution with the off-diagonal elements to be non-zero. To build the neural network model (presented in Fig. 6) with variational inference we use the "DenseVariational" function along with the input shape, followed by a batch normalization of the input layer. Instead of finding a single set of optimal weights, it learns a distribution over the weights that best describe the data. We use 8 neurons in the intermediate "DenseVariational" layer with the "sigmoid" activation function. Typically, the output is modeled as a mixture of gaussians, each having a mean and variance. Maintaining the input data from the test data constant and randomly sampling the weights from the distributions, at different number of iterations, we determine the uncertainties associated within the data and a specific model instantiation at each iteration with minimization of the negative log-likelihood loss.

We discuss the details of the probabilistic Bayesian ML model, its background, especially the implementation of the variational inference, the hyperparameters used in section 4 and corresponding model summaries in Fig. S3 of SI.

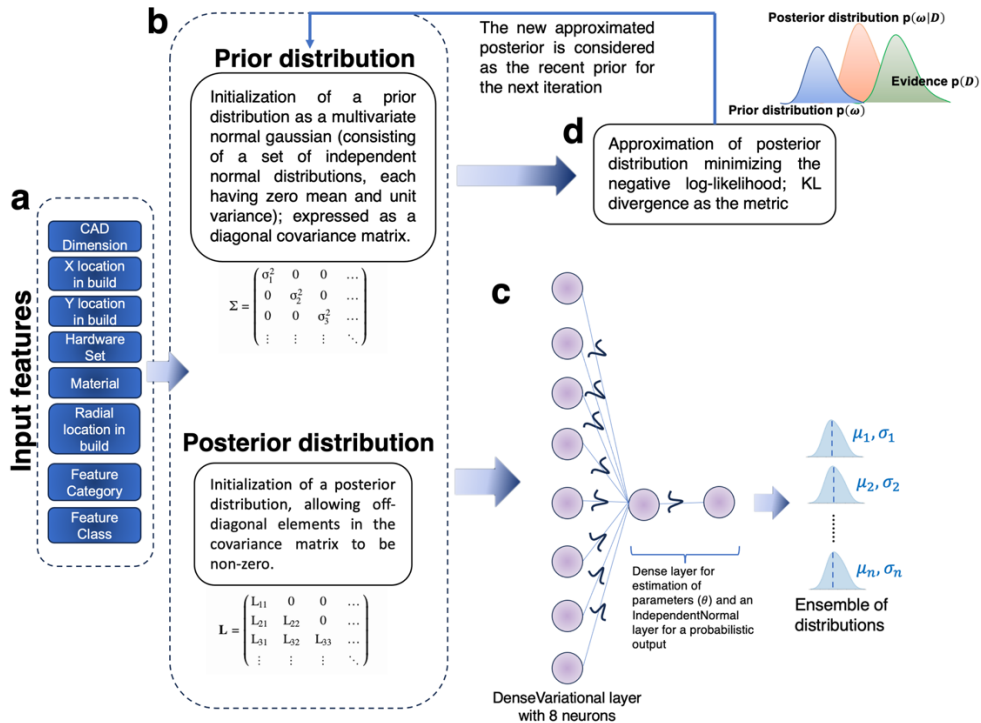

**Fig 6: Bayesian neural network working as an ensemble. a** Eight features were used as inputs to the probabilistic regression method, out of which six input features were manufacturing parameters, and rest of the features were measured feature descriptors. **b** Initially we define a the prior (consists of a set of gaussians with zero mean and variance) to define a posterior (a covariance matrix having off-diagonal elements non-zero) distribution. **c** Utilization of probabilistic layers for a probabilistic output, yields a set of distributions having unique means and variances. **d** The variational inference utilized in this work approximates a posterior distribution and maintains proximity with the prior distribution while minimizing the negative log-likelihood loss. The newly approximated posterior is considered as the recent prior for the next iteration.

## 3. Results and Discussions

In our work, we employ two distinct machine learning algorithms for prediction of Differences from Target (DFT) values. Deterministic ML models provide point estimates, yielding a single predicted value for each input instance. These models are computationally efficient and straightforward to interpret, making them valuable for scenarios where precise point predictions are the primary objective. In contrast, probabilistic ML models output probability distributions rather than a single-point estimate. This approach offers a more comprehensive representation of uncertainty, enabling the quantification of prediction intervals and aiding in decision-making under uncertainty. By employing both deterministic and probabilistic models, we aim to leverage their complementary strengths to enhance the accuracy and reliability of DFT predictions.

### 3.1 Deterministic ML algorithms

In the initial investigation in Fig. 7, we evaluate seven different deterministic machine learning algorithms. We found all of these algorithms possess the capability to predict feature DFT values. However, nonlinear models tend to exhibit superior performance compared to linear models, even when the training sample fractions are low. The present study illustrates the most optimal outcomes obtained from the Support Vector Regression (SVR) and Extreme Gradient Boosting (XGBoost) methods since they exhibit comparable levels of performance. Support Vector Regression (SVR) is a commonly used machine learning technique known for its extensive use. In Fig 7(b), we represent the average performance of the SVR model with corresponding training and testing sampling fractions, similarly presented in McGregor et. al. [16]. However, in Fig 7(c), XGBoost, a relatively recent and scalable machine learning system for tree boosting, has demonstrated nearly identical performance with Support Vector Regression. Fig 7(b) illustrates the performance of the SVR model in terms of training and testing, employing varying sampling fractions ranging from 10% to 90%. The data points in the graph represent the average root mean squared error values, the average performance of the model across 50 iterations. The bands in the graph indicate the standard deviation. We evaluate the accuracy, quantified by calculating the root mean square error ($RMSE$) between the predicted feature geometries using feature DFT and the corresponding measured feature geometries for the withheld testing data. This metric indicates the model's ability to accurately predict feature geometries. Training performance, also known as fit error, is quantified by the root mean square error ($RMSE$) between the model's predictions and the corresponding measurements obtained from the training data. This metric serves as an indicator of how effectively the model can capture the patterns and characteristics inherent in the training dataset. The models that exhibit superior performance are characterized by having the lowest testing $RMSE$, while their training accuracy is either comparable or somewhat smaller than the test accuracy. Similarly, as presented in reference [16], the SVR model demonstrates a remarkable ability to minimize prediction errors, reaching values approximately as low as 53 $\mu m$ (refer to Table 1). This level of accuracy is close to the process repeatability and significantly outperforms the standard deviation of the data, which stands at 180 μm. The anticipated optimal performance at 50 μm, is determined by considering the repeatability of the manufacturing process and the uncertainty in measurement, reported by the manufacturer itself [16]. The estimated value of 50 µm is in proximity to the asymptotic threshold of the performance of SVR.

Fig 7(b) illustrates that the predictive accuracy of SVR remains high even when using small sampling fractions, such as 25%, indicating its exceptional efficiency in utilizing data. As an illustration, the models that were trained to utilize a sampling fraction of 25% can predict the geometry of the remaining components, yielding an *RMSE* of around 65 *µm*. At smaller sampling fractions, the inclusion of additional training data yields more meaningful enhancements in prediction accuracy, but at higher sampling fractions, the gains in prediction error are comparatively less significant. Insufficient data during model training might be the probable source, as the performances of training and testing converge when higher sampling fractions are utilized. Furthermore, in Fig. 7(d) we assess the agreement between the predictions and measured values of DFT for the SVR model with the measured values for DFT, achieving the lowest *RMSE* value of 0.0529 mm.

**Table 1:** Comparison of performance of different deterministic ML models on test dataset (units of all the error metrics are in microns (mm))

| Algorithm | Average *RMSE* | Maximum *RMSE* | Minimum *RMSE* | Standard Deviation | Prediction Range |
|---|---|---|---|---|---|
| SVR | 0.05290 | 0.06991 | 0.04608 | 0.00559 | 0.01383 |
| kNN | 0.05826 | 0.06569 | 0.05031 | 0.00580 | 0.01538 |
| Decision Tree | 0.07123 | 0.08164 | 0.06231 | 0.00689 | 0.01933 |
| Random Forest | 0.05925 | 0.06380 | 0.05558 | 0.00390 | 0.00822 |
| GBM | 0.05615 | 0.06241 | 0.04878 | 0.00549 | 0.01363 |
| XGBoost | 0.05367 | 0.07451 | 0.04665 | 0.00736 | 0.01786 |
| LGBM | 0.05386 | 0.06590 | 0.04264 | 0.00474 | 0.01204 |
| MLP | 0.05852 | 0.05108 | 0.04631 | 0.00654 | 0.00477 |

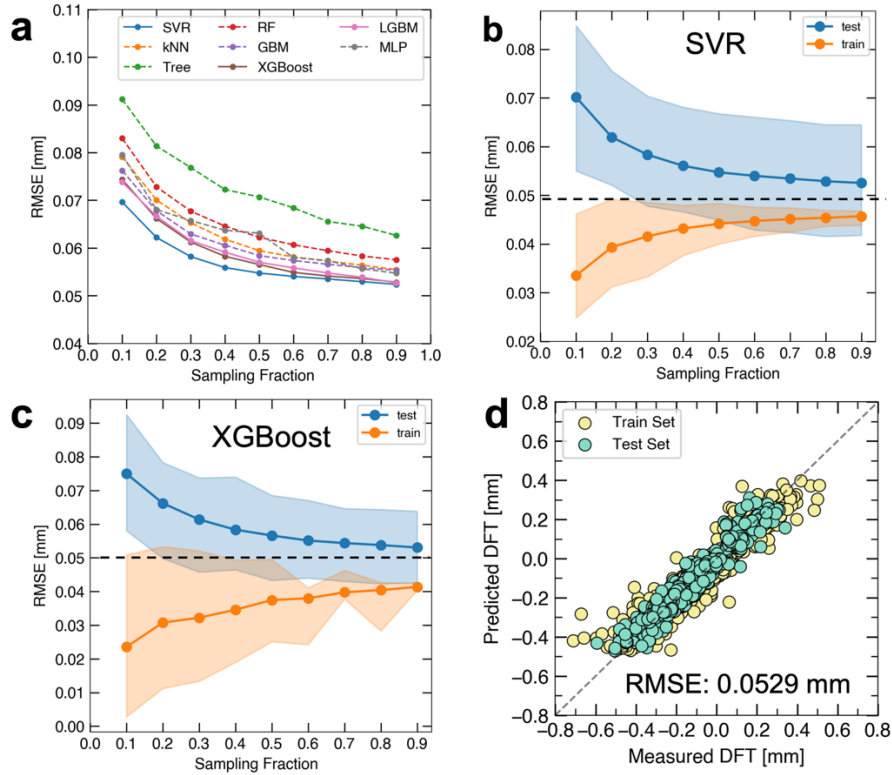

**Fig 7**: **Comparison of predictive capabilities of the different deterministic ML models. a)** Comparison of deterministic ML algorithms at different sampling fractions. The RMSE values across different sampling fractions are reported on the test dataset. **b** Performance of Support Vector Regressor (SVR) and **c** Xtreme Gradient Boosting (XGBoost) across various testing fractions. The points denote the average performance achieved in training and test across 50 iterations with corresponding standard deviations as error bands. d Parity plot between measured DFT vs. DFT predictions for support vector regression as the best performing regression method.

## 3.2 Probabilistic ML algorithms

We employ two distinct probabilistic machine learning approaches: Gaussian Process Regression (GPR) and Probabilistic Bayesian Neural Networks (BNNs) for our work. Probabilistic BNNs offer a flexible framework for modeling uncertainty by providing probability distributions over model parameters, allowing for robust uncertainty quantification. On the other hand, Gaussian Process Regression leverages a non-parametric approach [36] and offers a powerful tool for modeling uncertainty and making predictions. Our study centers on the utilization of probabilistic models to examine the projected predictions of DFT values while considering the accompanying uncertainties that deterministic models cannot offer.

### 3.2.1 Gaussian Process Regression (GPR)

In contrast to conventional regression models that yield a singular point estimate, a Gaussian Process (GP) regression model offers a probability distribution encompassing potential functions that may elucidate the data. For our study, we use the Scikit-Learn implementation of GPs that make use of prior over functions (usually assumed to be a zero-mean Gaussian process), updated based on observed data to form a posterior distribution. We start by splitting the dataset into 80:20 ratios and given a set of input points, GP regression allows us to compute a predictive distribution over the output values. This distribution gives us not only a mean prediction but also uncertainty estimates in terms of standard deviation. In Fig 8, we provide a comparison between the mean predictions using GP regression and ground truth from the test dataset for 50 test samples. In Fig 8, we represent a parity plot between the mean predictions and measured values with corresponding aleatoric uncertainty for each data point from the mean predictions. We present the uncertainties from the mean predictions of the DFT values with corresponding standard deviations at 95% confidence. As we evaluate the GP regression model, we get the test RMSE of 49$\mu m$, which is accurate than acceptable RMSE of 50 $\mu m$. Moreover, it is worth noting that while the Gaussian Process Regression (GPR) can estimate the noise variance inherent in the data, that represents aleatoric uncertainty, it does not offer insights into model uncertainty [36]. To model the epistemic uncertainty, there is a typical need to incorporate additional techniques such as the Bayesian methods or ensemble methods to account for model uncertainty or parameter uncertainty.

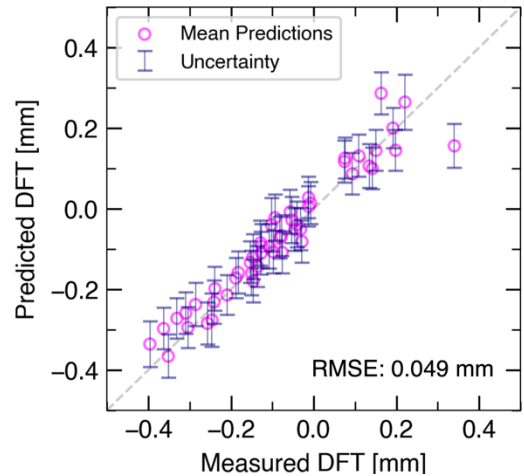

**Fig 8**: Parity plot between measured DFT vs. DFT mean predictions with corresponding aleatoric uncertainty plotted as error bars for each mean prediction from the GP Regressor model.

### 3.2.2 Bayesian Neural Networks (BNN)

We train a Bayesian neural network using the experimental training data for the fabrication and measurement of produced parts using Digital Light Synthesis (DLS) and utilize it for dimensional accuracy predictions on unseen data. The predictions obtained from the test set, together with their corresponding uncertainties, are compared with the measured values that were estimated using experiments. A Bayesian neural network is distinguished by its probability distribution across weights (parameters) and/or outputs. The code implementation of a Bayesian neural network exhibits modest variations depending on whether aleatoric, epistemic, or both uncertainties are considered. In our work, we explore two different approaches for using the Bayesian neural network, where in the first case that characterize only aleatoric uncertainty and in the other case, can separate both aleatoric and epistemic uncertainty.

### 3.2.2.1 Preliminary approach – BNN with trainable mean and variance

The objective of our preliminary approach is to account for the inherent noise present in the data. To achieve this, we propose training a regression model that yields a normal distribution with trainable mean and variance, a single output distribution. This approach significantly differs from deterministic regression model and point estimation methods. The model's mean and standard deviation are trainable parameters, allowing for flexibility in capturing the variability of the data. To represent a normal distribution based on the output generated by the fully-connected layers, we employ a probabilistic layer that yields a distribution object as output. Following the completion of training, we extract 50 test samples and provide a representation displaying the parity between the mean predictions and measured values of DFT with corresponding standard deviation as the aleatoric uncertainty in Fig 9.

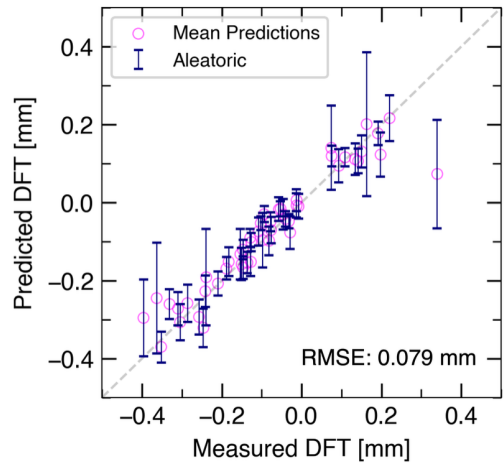

**Fig 9**: Parity plot between measured DFT vs. DFT predictions with corresponding aleatoric uncertainty plotted as error bars for each mean prediction from 50 test data samples with 95% confidence intervals for each mean prediction for the DFT values.

### 3.2.2.2 Probabilistic approach – BNN working as an ensemble of networks

On the other hand, in our second approach, we consider evaluation of both aleatoric and epistemic uncertainty using a probabilistic Bayesian NN model. In the context of probabilistic analysis discussed in this study, we construct the neural network by combining two elements - a network architecture and prior distribution over the weights. The selected architecture comprises the "DenseVariational" layer, containing 8 units that employ the variational inference method to estimate a surrogate posterior distribution for both the kernel matrix and the bias terms. This architecture connects 8 input features to a single output unit and employs the non-linear "sigmoid" activation function. We consider the initial probability density function for the weights to be a conventional normal distribution. This particular model possesses a greater number of parameters since each weight is characterized by a normal distribution with distinct mean and standard deviation values, hence increasing the parameter weight counts. We resample the weights by iterating multiple times to provide diverse predictions while keeping the input data constant (without shuffling), therefore causing the neural network model to function as an ensemble. In Fig 9, we separate both the aleatoric and epistemic uncertainties along with the mean predictions from 50 test samples and measured DFT values. We assess the uncertainties, particularly the aleatoric uncertainty ($\sigma_{aleatoric}$), by taking the root mean squared of variances ($\sigma_i^2$) over 200 iterations from the ensemble of the output distributions. Next, we average over the standard deviations of the mean predictions ($\mu_i$) from 200 iterations to evaluate the epistemic uncertainty ($\sigma_{epistemic}$). Therefore, we obtain the uncertainty estimates as a mixture of Gaussians over $N$ iterations (from 200 iterations) and represent those in Eq. 2 and Eq. 3.

$$\sigma_{aleatoric} = \sqrt{\frac{1}{N}\sum_{i=1}^{N} \sigma_i^2} = \sqrt{mean(\sigma_i^2)} \qquad (2)$$

$$\sigma_{epistemic} = stdev(\mu_i) \qquad (3)$$

The model provides on-average correct predictions that do not systematically overestimate or underestimate the true values, while the estimated overall uncertainty closely matches the actual variability in the data. Also, from Fig. 10(a), we observe there are several datapoints (lies within the range of $\pm 0.4$ to $\pm 0.6$ mm) of the predicted and measured DFT values from both the training and test set that works as outliers and gives a "sigmoid" shape to the predictions. In Fig. 10(b), we observe the error bars of the mean predictions in the parity plot touch the best-fit line, signifying the model's consistency in accuracy and bias. This alignment between the model's mean predictions and the parity line indicates reliable and unbiased performance, making it a favorable choice as the probabilistic predictive model we use in this work.

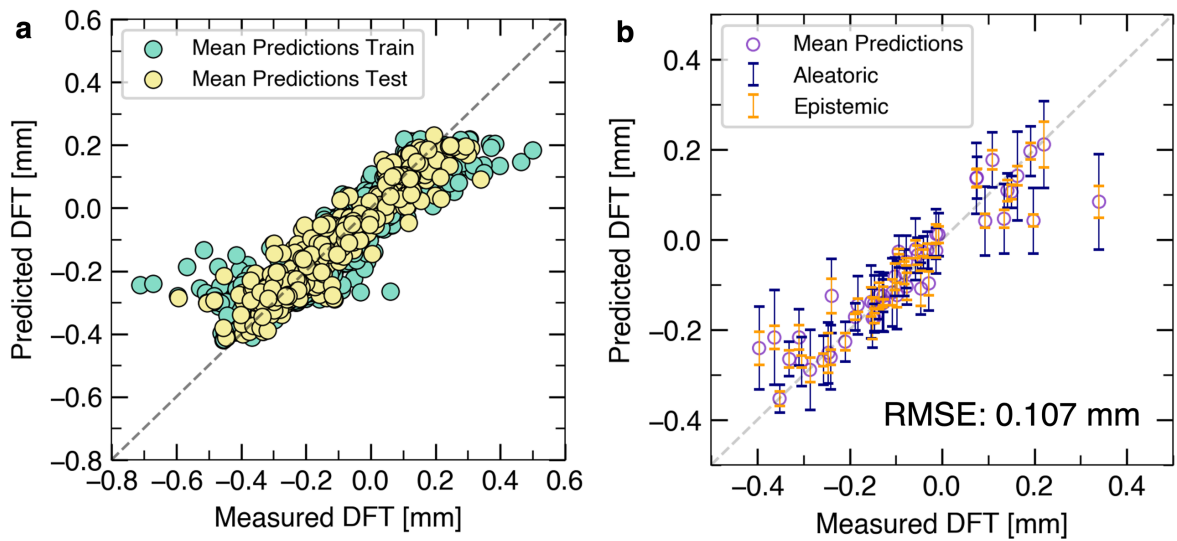

**Fig 10**: **Parity plot and uncertainty quantification from DFT predictions while considering BNN working as an ensemble of networks. a)** Sigmoid shape distribution of the entire data while we plot the predicted values against the true values both from the training and the test set. **b)** Parity plot between measured DFT vs. DFT predictions with corresponding aleatoric and epistemic uncertainties plotted as error bars with 95% confidence intervals for each mean prediction of DFT values on test set.

It is noteworthy that, the predictive accuracy of our preliminary approach is superior when compared with our second approach, the probabilistic BNN. The probabilistic BNN yields a predictive accuracy or the test RMSE of 0.107 mm, while the preliminary Bayesian neural network model with trainable mean and variance yields a lesser test RMSE value which is 0.079 mm. However, it is worth noting that the probabilistic BNN model is successful in quantification of both aleatoric (±63.96 μm) and epistemic uncertainty (±20.68 μm), while the preliminary approach only characterizes aleatoric uncertainty (±54 μm). Additionally, the performance difference between the two Bayesian neural network approaches may stem from factors such as differences in model complexity, sensitivity towards hyperparameter tuning, the size of the ensemble in DenseVariational layers, training duration, and the adequacy of posterior sampling. Experimenting with adjustments to these aspects could help reconcile the performance gap and enhance the probabilistic approach's accuracy on the test set.

Fig 11 illustrates the predictive capabilities of the network ensemble regarding the mean and variance predictions of the difference from the target (DFT) when a substantial amount of data (1215 training data points, equivalent to 80% of train data split) is employed. We observe that with the decrease in training data points, the variations in the posterior distributions are larger, strongly specifying the presence of epistemic uncertainty within the trained model. Fig 11 (d to f) provides more evidence that the anticipated epistemic uncertainty effectively diminishes with an increase in the quantity of training data, namely from 10% to 50% to 80% training percentage.

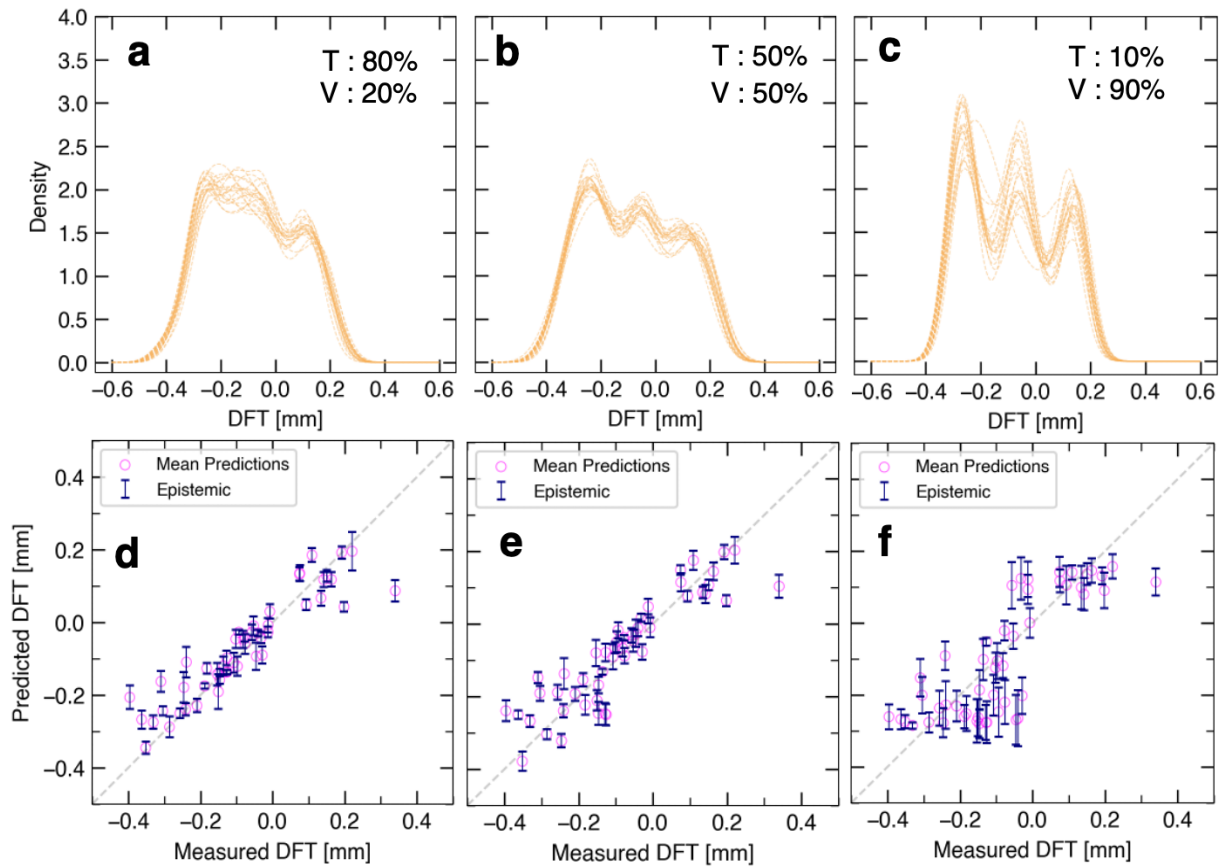

**Fig 11: Variations in the posterior distribution samples and quantified epistemic uncertainty with varying ratios of training and validation data. a, b, c** illustrates the predictive capabilities of the probabilistic network ensemble in terms of the posterior probability distributions of the difference from target (DFT). **d, e, f** parity plots between the measured and predicted DFT values illustrating the variations in aleatoric and epistemic uncertainties for the train-test

split ratios. In this observation, it appears that the epistemic uncertainty remains less while considering reasonable ratios of training and testing data splits, such as 50% and 80% for training data. As we decrease the training data to 10%, we find the error bars on the mean predictions are prominent, hence confirming the anticipated increase in epistemic uncertainty.

To provide a conclusive verification, in section 4 of SI, in Fig. S4, the trends in aleatoric and epistemic uncertainty for the train-validation split ratios. We split the training dataset in different ratios ranging from a 10% training split to a 99% training split. Here, we observe that the aleatoric uncertainty exhibits a decreasing pattern with intermittent increase while considering appropriate ratios of training fractions, such as 50% and 80% for the training data. We observe a consistent downward trajectory in epistemic uncertainty in conjunction with an increase in the number of training data points.

Besides aleatoric uncertainty, quantification of epistemic uncertainty using a machine learning model is critically important as it enables robust decision-making by providing a measure of the model's confidence and acknowledging areas where it lacks knowledge. Selecting between a Probabilistic Bayesian Neural Network (BNN) and Gaussian Process Regression (GPR) hinges on the nuanced trade-offs in our analysis. Despite the BNN achieving lower accuracy or higher RMSE value, its unique strength lies in its capacity to capture both aleatoric and epistemic uncertainties. This is particularly valuable in applications like additive manufacturing, where comprehending and quantifying uncertainty is paramount. The BNN's probabilistic nature equips it to provide richer insights into the nature of uncertainties, enhancing robust decision-making under ambiguous circumstances. However, GPR, being more interpretable and computationally efficient, might be preferred in cases where model interpretability or computational resources are limited. In essence, the choice between the two models can be dictated by the analytical needs, balancing predictive accuracy, interpretability, computational constraints, and the necessity of uncertainty characterization in the problem statement.

## 4. Conclusions

The current investigation centers on an estimation of dimensions related to the geometry of additively manufactured components. To emulate the additive manufacturing (AM) production process, we utilize an experimental dataset comprising 405 parts produced in nine distinct experiments [16]. Using two different machines, three polymer materials, and two-part design configurations in each run, McGregor et al. [16] analyzed three distinct part designs, measuring five crucial features for each. This yielded a total of 2025 feature measurements. Data models are employed to represent design details and manufacturing conditions, encompassing both continuous and categorical factors. Our work initiates with utilizing two distinct machine learning approaches to make predictions on Differences from Target (DFT) values. Deterministic machine learning models provide point estimates, resulting in a singular projected value for each input instance. In contrast, probabilistic machine learning models generate probability distributions as outputs instead of providing a singular point estimate.

While we employ deterministic ML algorithms, we utilize an 80:20 train-test data split with dual Monte Carlo subsampling [] and nested k-fold cross-validation. We employ seven non-linear deterministic ML models trained on a randomly selected subset of 405 parts, accurately predicting the geometric properties of unsampled parts. The Support Vector Regression (SVR) model better accuracy amongst other deterministic methods; around 53 µm (also reported in ref [27]), close to the manufacturer's process repeatability, and significantly smaller than the standard deviation (180 µm) of the data. Considering an expected ideal performance at 50 µm, we account for manufacturing process repeatability and measurement uncertainty [37]. This value aligns with the SVR regressor's asymptotic threshold.

We utilize two distinct probabilistic machine learning methodologies: Gaussian Process Regression (GPR) and Probabilistic Bayesian Neural Networks (BNNs). First, we use the Gaussian Process Regression (GPR) model for feature geometry predictions, which estimates the noise variance inherent in the data but fails to provide insights about the epistemic uncertainty. Moving forward, we use the Probabilistic Bayesian neural networks (BNNs), which present a versatile framework for representing uncertainty through the provision of probability distributions over model parameters that enable robust quantification of uncertainty. We use two different approaches to develop the Bayesian NN models. The initial approach addresses the intrinsic noise that is inherent in the dataset, the aleatoric uncertainty. The mean and standard deviation of the model are trainable parameters, enabling the model to effectively capture the variability present in the data. In contrast, our second method involves in the assessment of both uncertainties through the utilization of a probabilistic Bayesian neural network model. The iterative resampling of the weights of the probabilistic BNN architecture creates diverse predictions, transforming the neural network into an ensemble. Our second approach, the probabilistic BNN demonstrates low accuracy compared to our preliminary approach in predicting feature geometry dimensions, achieving an accuracy of around 107 µm, while our preliminary approach achieves an accuracy of around 79 µm. However, it is worth noting that the probabilistic BNN model with the higher accuracy value quantifies both the aleatoric ($\pm 63.96$ µm) and epistemic ($\pm 20.68$ µm) uncertainties.

Analyzing an additive manufacturing (AM) dataset through the lens of a Probabilistic Bayesian Neural Network (BNN) and the concurrent quantification of both epistemic and aleatoric uncertainties provides a robust foundational platform for an array of promising avenues of research. As a prospective extension of this work, the refinement of uncertainty modeling within the BNN through the implementation of advanced Bayesian methodologies [30], holds the potential for more precise quantification of uncertainties, specifically in AM applications. The microstructure influences the strength of an additively manufactured component and tailored microstructures serves as a critical determinant in material properties, performance, and quality. BNNs provide a probabilistic framework that accounts for the quantification of uncertainty associated with microstructure predictions. In AM processes, where variations in materials composition, cooling rates, and printing parameters can lead to significant microstructural variability, accurate assessment of uncertainty is crucial. BNNs enable the trustworthy estimation of uncertainty intervals for predicted microstructural features while providing valuable and deeper insights towards the reliability and confidence of these predictions. The ability of accurate prediction of microstructural features using BNNs can facilitate the optimization of process parameters and can play a crucial role in materials selection in AM. By leveraging uncertainty quantification, it is more transparent towards identification of optimal printing conditions that minimize defects, such as porosity, grain boundary misalignment, or residual stresses, while maximizing desired microstructural characteristics, such as grain size, phase distribution, or mechanical properties, while adversely helping towards the enhancement of the overall quality and performance of additively manufactured components. Repeatability in experimental outcomes is a crucial factor while identifying anomalies and defects in an additive manufacturing (AM) setup, even when the processing conditions remain constant. Maintaining nearly identical results under invariant conditions enables swift detection of any deviations from the anticipated outcome. Probabilistic ML models are a trustworthy solution to replicate those systematic deviations. For instance, in such cases, employing BNNs with a simple averaging approach across repeated experiments can help capturing the significant deviations from the expected result or uncertainties. Moreover, machine learning models, with relevant microstructural features, trained on simulated data can be relevant surrogates for systematic exploration of different processing conditions relevant to additive manufacturing experiments and mapping the relationships between process, structure and materials properties. Often times the simulation methods are stochastic (i.e.- Kinetic Monte Carlo, Cellular Automata, Phase-field methods), thus using a probabilistic ML model instead of a deterministic one can help circumvent the issue of stochasticity while mapping between process, structure and/or properties [14,15].

Furthermore, there is significant scope for the development of real-time monitoring of data and uncertainty evaluation [37] aiming to ensure strict adherence to established quality standards during the additive manufacturing processes. Furthermore, there are potential avenues for research in materials science to develop correlations between uncertainty and material properties, thereby generating significant insights into the additive manufacturing process. We consider the development of interpretability techniques to enhance the trustworthiness of the probabilistic regression methods and adoption in complex industrial settings is a critical facet of future work, as is the exploration of data augmentation strategies and transfer learning methodologies to enhance the adaptability of the probabilistic regression methods in diverse manufacturing environments.

## Declaration of interests

The authors declare that they have no known competing financial interests or personal relationships that could have appeared to influence the work reported in this paper.

## Code and data availability

The code for the deterministic ML models, the probabilistic ML models and the dataset from ref. [27], that we use to perform the uncertainty quantification calculations as implemented in the present study is publicly available (https://github.com/dipayan-s/UQ-Repository).

## Acknowledgements

This work performed at the Center for Nanoscale Materials, a U.S. Department of Energy Office of Science User Facility, was supported by the U.S. DOE, Office of Basic Energy Sciences, under Contract No. DE-AC02-06CH11357. This material is based on work supported by the DOE, Office of Science, BES Data, Artificial Intelligence, and Machine Learning at DOE Scientific User Facilities program (ML-Exchange and Digital Twins). SKRS would also like to acknowledge the support from the UIC faculty start-up fund. All authors thank Suvo Banik and Partha Sarathi Dutta for their useful comments and discussion during the research activity. This research used resources of the National Energy Research Scientific Computing Center (NERSC), a US Department of Energy Office of Science User Facility located at Lawrence Berkeley National Laboratory, operated under Contract No. DE-AC02-05CH11231.## Authorship contribution statement

DS, AC and SKRS conceived the project. DS developed the code and performed the uncertainty quantification calculations with inputs from AC. All the authors provided feedback on the workflow. DS, AC and SKRS wrote the manuscript with inputs from all co-authors. HC, SM provided feedback on the manuscript. All authors participated in the discussion of results and provided useful comments and suggestions on the various sections of the manuscript. SKRS supervised the overall project and research activity.

## References

[1] MacGregor, J.F. and Kourti, T., 1995. Statistical process control of multivariate processes. *Control engineering practice*, *3*(3), pp.403-414.

[2] Fisher, R.A., 1936. Design of experiments. *British Medical Journal*, *1*(3923), p.554.

[3] Mun, J., 2006. *Modeling risk: Applying Monte Carlo simulation, real options analysis, forecasting, and optimization techniques* (Vol. 347). John Wiley & Sons.

[4] Chase, K.W. and Parkinson, A.R., 1991. A survey of research in the application of tolerance analysis to the design of mechanical assemblies. *Research in Engineering design*, *3*(1), pp.23-37.

[5] Schroeder, R.G., Linderman, K., Liedtke, C. and Choo, A.S., 2008. Six Sigma: Definition and underlying theory. *Journal of operations Management*, *26*(4), pp.536-554.

[6] Huang, D.J. and Li, H., 2021. A machine learning guided investigation of quality repeatability in metal laser powder bed fusion additive manufacturing. *Materials & Design*, *203*, p.109606.

[7] Liu, J., Ye, J., Silva Izquierdo, D., Vinel, A., Shamsaei, N. and Shao, S., 2022. A review of machine learning techniques for process and performance optimization in laser beam powder bed fusion additive manufacturing. *Journal of Intelligent Manufacturing*, pp.1-27.

[8] Song, L., Huang, W., Han, X. and Mazumder, J., 2016. Real-time composition monitoring using support vector regression of laser-induced plasma for laser additive manufacturing. *IEEE Transactions on Industrial Electronics*, *64*(1), pp.633-642.

[9] Wang, Q., Song, L., Zhao, J., Wang, H., Dong, L., Wang, X. and Yang, Q., 2023. Application of the gradient boosting decision tree in the online prediction of rolling force in hot rolling. *The International Journal of Advanced Manufacturing Technology*, *125*(1-2), pp.387-397.23

# Supplementary Materials

## 1. Dual Monte Carlo Subsampling method

Fig. S1 represents the training and testing method, that use a dual Monte Carlo subsampling [1] methodology with nested k-fold cross-validation, tied with hyperparameter tuning for training and evaluation of the models, and reduce bias and overfitting. The proposed strategy leverages on an unbiased hyperparameter tuning procedure and predictions with accurate outcomes [2]. The model is tested on data without exposure to the training procedure and the withheld data, that is not used for training and testing either by the ML model, signifies the robustness of measurement in the experimental conditions where generation and collection of data is often costly, and thus enabling measure of data efficiency.

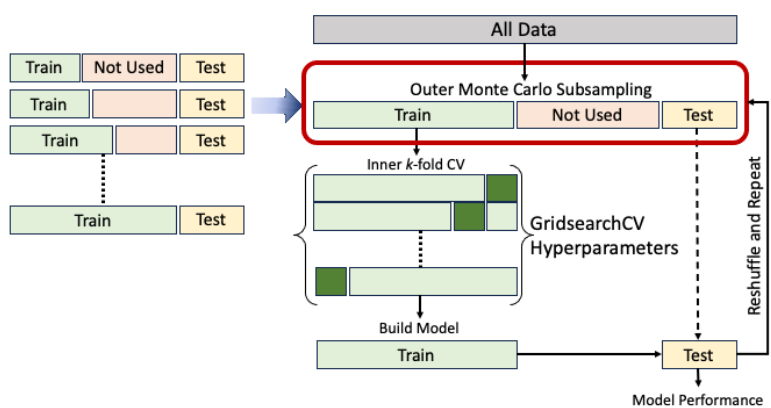

**Fig. S1:** The proposed approach involves the utilization of dual Monte Carlo subsampling in conjunction with a stacked k-fold cross-validation (3 folds) pipeline for the training and testing of machine learning models. Within the outer loop, the data undergoes a random splitting process, resulting in the creation of distinct sets for training, testing, and unused data. The data is rearranged, and this procedure is iterated (here we used 50 iterations) for each iteration of the outer loop. The utilization of the Monte Carlo method offers a sound approximation for the performance of the model.

The dual Monte Carlo subsampling method constitute two nested loops and subsequent steps for data pre-processing, subsampling, scaling, hyperparameter tuning, training, testing at a fixed number of iterations. In the outer loop, random sampling of data was performed to split data into training, test, and sometimes unused sets. The training data was normalized using the Minmax scaling method to have mean zero and a standard deviation of unity. We use the standardized training data for hyperparameter tuning by leveraging k-fold cross-validation and grid search. In the outer loop, we define the settings for the tuned hyperparameters, and training data trained using a single model instance, and the withheld test data for evaluation of the model. Next, we assess the model performance using the withheld testing data and represent it as root mean square error ($RMSE$) between the measurement of the part geometries and geometry predictions. Using two nested loops, where the data is randomly shuffled at each stage, it is possible to execute multiple iterations of the train-test procedure. Each iteration trains a distinct and unique model based on training data and evaluates it based on withheld testing data. In Table S1 we elaboratively discuss the details of each input feature and the target variable difference from target (DFT), the key metric to account for the dimensional accuracy of the fabricated parts.

**Table S1:** Detailed description of the input features and the target variable

| Input feature | Description | Type |
|---|---|---|
| *Manufacturing parameters* | | |
| Hardware set | Machine number used for manufacturing of the part geometry (2 Carbon M2 printers were used) (Parameter states: 1, 2) | Categorical |
| Material | 3 different materials were used to fabricate the parts namely, UMA, RPU and EPX. | Categorical |
| Thermal cure | The materials used were ultraviolet curable photopolymers; followed by an oven bake for cross-linking. | Categorical |
| Layout | 2 different layouts were used to show arrangements of different part designs, within the print area. | Categorical |
| x-coordinate | Cartesian co-ordinate location of the fabricated part within the build area. | Continuous |
| y-coordinate | Cartesian co-ordinate location of the fabricated part within the build area. | Continuous |
| R-coordinate | Radial co-ordinate location of the fabricated part within the build area. | Continuous |
| Unique build ID | Build ID identified the exact build in which a part was manufactured. A total of 405 parts were built across 9 different build ids with 5 measurements for each part. (example: build ids 1 through 9) | Categorical |
| *Measured feature descriptors* | | |
| Part design | 3 different parts designs were fabricated, namely, clip, plug and bracket. | Categorical |
| Nominal dimension | Measured feature dimensions from experiments and reference geometry dimensions from computer-aided design. | Continuous |
| Feature class | The measured dimensional entity of the manufactured part. (example: diameter, length, thickness, height) | Categorical |
| Feature category | Descriptor for the feature class that describe the topology of the geometry and categorize them. (example: inner/outer diameter) | Categorical |
| Unique feature ID | Feature ID identified the unique feature of the part, such as bracket_dist_mm_c, which occurred once for a bracket design for a unique build id and present for each fabricated bracket. | Categorical |

| Target variable | Description | Type |
|---|---|---|
| Difference from Target (DFT) | Difference between the measured dimension (i.e.- length, inner/outer diameter of a bracket etc.) and the refence CAD geometry. | Continuous |

# 2. Implementation of Deterministic ML algorithms:

## 2.2.1 k-Nearest Neighbors (kNN):

k-NN [3] is a straightforward and intuitive machine learning algorithm used for classification and regression tasks. Being non-parametric, it makes no assumptions on the distribution of the underlying data. Instead, it makes predictions based on the data itself. The training data points, and their related labels or values are simply memorized by the algorithm. In regression tasks, the technique computes the average (or weighted average) of the target values of the 'k' nearest neighbors rather than assigning a class label. This average becomes the new data point's anticipated value. The most crucial kNN parameter is 'k', which denotes the number of neighbors to consider while making predictions. The algorithm's performance is influenced by the selection of 'k'. A prediction made with a small "k" might be noisy, whereas one made with a greater "k" value might be overly broad. The effectiveness of the algorithm is highly dependent on the selection of the distance metric. Euclidean distance, Manhattan distance (measured in city blocks), and cosine similarity are examples of common distance measurements. The distance metric establishes the "closeness" of the data points as determined by the algorithm.

## 2.2.2 Support Vector Regression (SVR):

Support vector machines (SVM) [4] are one of the most prevalent classification algorithms. SVMs locate "hyperplane" decision boundaries, that divide classes in high-dimensional space using a linear or kernel function. These hyperplanes are produced by positioning the hyperplane at the farthest distance possible between the extreme data points or the support vectors, which provides maximum margin. Regression problems can be addressed with SVMs. The model then anticipates continuous values as the target variable instead of class labels.

## 2.2.3 Decision Tree Regression (Tree):

An additional well-liked subset of ML algorithms is decision trees [5]. As a quality check, data is constantly divided into nodes using a splitting rule (depending on locally optimal choices that minimize the mean squared error) until further splitting neither improves the model prediction nor reaches a predetermined depth of the tree. In contrast to most ML models with scale variance, this allows for very straightforward and understandable predictions and requires minimal data preparation (no feature normalization or standardization is necessary). If minor data variations lead to unstable and erroneous tree constructs, there is a problem. As a result, after training individual trees, there is frequently considerable variance, and when there is an imbalance in the data, individual trees tend to overfit the input data. By incorporating various tree architectures, ensemble approaches, like a random forest classifier or regressor, might introduce some randomization to mitigate such issues. The variance is decreased, and an overall more accurate model is produced by averaging the predictions of each tree.

*2.2.4 Random Forest (RF):*

A well-liked ensemble learning approach for machine learning is Random Forest [6] which combines the predictive power of many decision trees. To avoid overfitting and improve forecast accuracy, it works by training many decision trees on random subsets of the data and features. It is an invaluable tool for classification and regression tasks across multiple areas, from banking and healthcare to image analysis and recommendation systems, by pooling the predictions of multiple trees and producing robust and extremely accurate results. It is a flexible option for real-world machine learning applications due to its capacity for handling diverse data types, feature importance ranking, and resistance to overfitting.

*2.2.5 Gradient Boosting Machine (GBM):*

Gradient Boosting [7] is a powerful machine learning algorithm that sequentially builds an ensemble of decision trees to create a highly accurate predictive model. It works by iteratively improving the model's predictions by focusing on the errors made by previous trees. Gradient Boosting is known for its exceptional predictive performance and is widely used in various fields such as finance, healthcare, and natural language processing. It is particularly effective in handling complex, non-linear relationships in data and often outperforms other algorithms in predictive accuracy. However, it may require careful tuning of hyperparameters to prevent overfitting.

*2.2.6 Xtreme Gradient Boosting (XGBoost):*

Xtreme Gradient Boosting [8], termed as XGBoost, is an advanced and extremely efficient gradient boosting implementation. It is renowned for its speed and precision in machine learning applications. Regularization and parallel processing are utilized by XGBoost to prevent overfitting and substantially reduce training time. Its advantages include the ability to capture complex data relationships, speed, handling of missing data, and support for feature importance ranking. However, XGBoost requires careful hyperparameter tuning, lacks the interpretability of simpler models, and may not perform optimally on very small datasets, limiting its applicability in such cases.

*2.2.7 Multi-layer Perceptron (MLP):*

Artificial neural networks (ANNs) [3] have garnered significant interest in the field of machine learning due to their effectiveness in addressing various computational challenges. The multilayer perceptron (MLP) is a type of neural network that employs a forward-predicting approach and is trained iteratively on a given dataset with the objective of minimizing a specified loss function [3]. During each iteration, the trainable parameters, commonly referred to as weights, undergo updates. The regularization term is an additional component of the model's functionality that serves to mitigate the risk of overfitting the model to the training data during the computation of the loss. Moreover, Multilayer Perceptrons (MLPs) are widely regarded as one of the most intricate machine learning algorithms in terms of optimization due to their comparatively large number of hyperparameters.

**Table S2: Hyperparameters for deterministic machine learning models,** here we present the optimal hyperparameters utilized for the machine learning models using GridsearchCV tied inside the Dual Monte Carlo subsampling algorithm, here we use 5 folds of cross-validation of the data with negative root mean squared error as the metric during hyperparameter optimization.

| Deterministic ML Algorithms | Hyperparameters |
|---|---|
| KNN | 'metric': 'euclidean', 'n_neighbors': 6 |
| SVR | 'epsilon': 0.03, 'gamma': 'scale', 'kernel': 'rbf' |
| Decision Tree | 'max_depth': 20, 'min_samples_leaf': 5, 'criterion': "absolute_error" |
| Random Forest | 'bootstrap': True, 'max_features': 3,'min_samples_leaf': 3, 'n_estimators': 300 |
| Gradient Boosting | 'learning_rate': 0.3,'loss': 'squared_error','max_leaf_nodes': 30,'n_estimators': 120 |
| XGBoost | 'learning_rate': 0.1, 'max_depth': 5, 'n_estimators': 100,'subsample': 0.9 |
| LightGBM | 'learning_rate': 0.1,'max_depth': 7,'n_estimators': 150,'num_leaves': 31 |
| MLP | 'activation': 'tanh','hidden_layer_sizes': (16, 8, 4),'learning_rate': 'constant','max_iter': 5000,'solver': 'lbfgs'} |

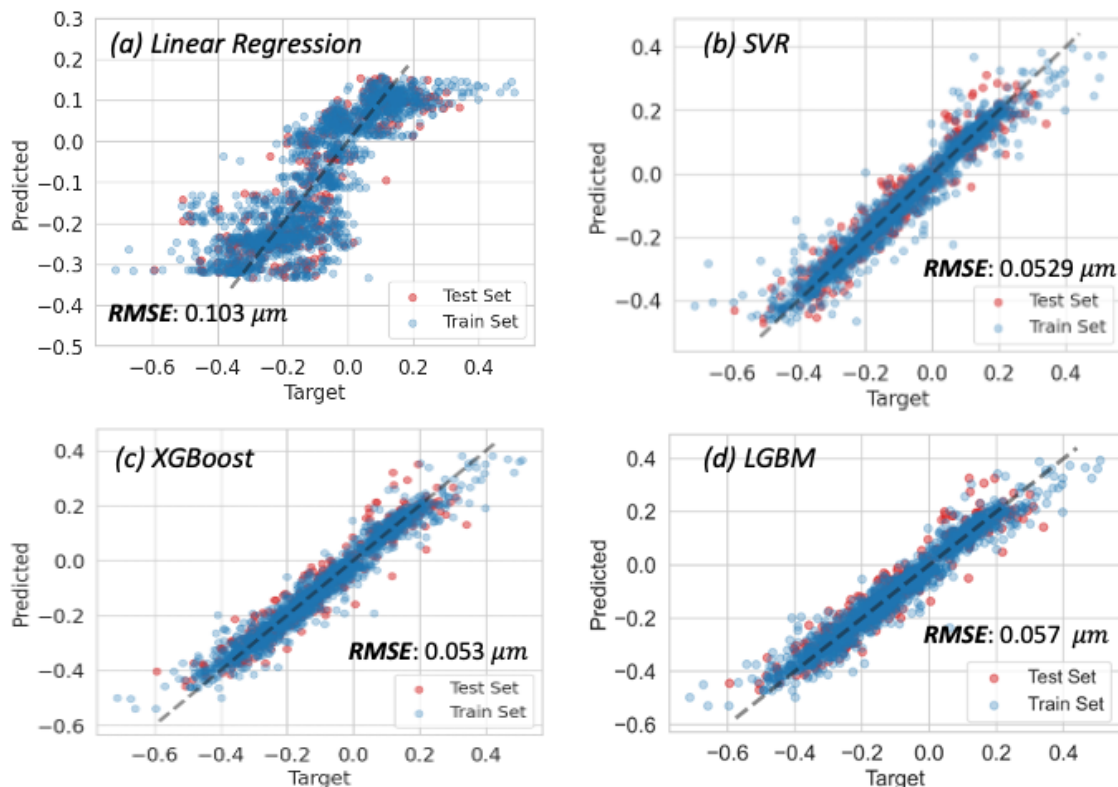

**Fig S2**: Parity plot between the measured and predicted DFT values as we implement different deterministic ML models (a) Linear Regression and rest of the three best performing deterministic ML models, (b) Support Vector Regression (c) Extreme Gradient Boosting, (d) Light Gradient Boosting, with their corresponding RMSE values. For each of the model implementation, specifically while we fit our data to linear regression, we observe a 'sigmoid' or a 'S' curve formation that provide subsequent evidence towards the presence of non-linearity in the data.

# 3. Implementation of Bayesian Neural Networks:

Given the mathematical representation of Eq. (1) in the main manuscript, and under the assumption that the training data points are distinct and identically dispersed, the likelihood of the entire dataset $\mathcal{D}$ is determined using the Eq (1) given below. The presence of noise in the dependent variables is not considered in this study, following the approach outlined in the reference provided [8].

$$p(\mathcal{D}|\omega) = \prod_{i=1}^{n} \mathcal{N}(y_i, f^{\omega}(x_i), \sigma_n^2) \tag{1}$$

Epistemic uncertainties, conversely, are assessed by constructing a probability model that encompasses the network weights, $\omega$ of the given model. In a Bayesian learning framework, a prior probability density function (pdf) of weights, is $p_0(\omega)$ is initialized in combination with the model parameters, The pdf gets updated utilizing the data and Bayes theorem after each feedforward mechanism to yield a posterior pdf $p(\omega|\mathcal{D})$, computed using Eq. (2):

$$p(\omega|\mathcal{D}) = \frac{p(\mathcal{D}|\omega)p_0(\omega)}{p(\mathcal{D})} \tag{2}$$

During the prediction phase, when presented with an input $(x^*)$, it is possible to calculate the predictive probability; $p(y^*|x^*, \mathcal{D})$ of the corresponding output given by:

$$p(y^*|x^*, \mathcal{D}) = \int p(y^*|x^*, \omega)\, p(\omega|\mathcal{D}) d\omega \tag{3}$$

From a classical perspective, as provided in reference [8], using the law of total expectation and total variance [8], the total uncertainty predictions are obtained as follows:

$$Var(y^*|x^*, \mathcal{D}) = \sigma_n(x^*)^2 + Var_{p(\omega|\mathcal{D})}(f^{\omega}(x^*)) \tag{4}$$

In the above equation, the term $\sigma_n(x^*)^2$ is the variance resulting from the inherent noise present within the data (aleatoric uncertainty) and the second term results from the epistemic uncertainty computed using variational inference.

However, the computation of an analytical solution for the posterior probability $p(\omega|\mathcal{D})$ in neural networks is infeasible due to the complex integral present within the weights. To address this challenge, researchers employ a range of approximate techniques, including the use of Variational Bayes [10]. The objective of this strategy is to effectively approximate the posterior distribution given the neural network weights. In the context of Variational Bayes, the posterior distribution is approximated using a secondary function referred to as the variational posterior. The variational inference works by approximation of a posterior distribution and the distribution is parameterized by a set of parameters, represented as q(ω|θ), acquired by the process of optimization. Here we consider the distribution to be a Gaussian distribution and given as [14],

$$q(\omega|\theta) = \prod_{i=1}^{n} \mathcal{N}(\omega_i, \mu_i, \sigma_i) \tag{5}$$

where, θ represents a set of parameters including the means and variances for each of the independent Gaussians; and $n$ is the total number of weights.

The primary concept is seeded around the selection of a functional form for the variational posterior, which can be optimized efficiently in order to closely resemble the genuine posterior. A frequently employed method for assessment of the accuracy of an approximation is by evaluating the Kullback-Leibler (KL) divergence between the variational posterior and the genuine posterior. The Kullback-Leibler divergence is a metric used to quantify the dissimilarity between two probability distributions, with a value of zero indicating that the two distributions are identical. The KL divergence between the variational posterior, $q(\omega|\theta)$, and the actual posterior, $P(\omega|\mathcal{D})$, can be defined based on the observed data, $\mathcal{D}$ [14].

$$D_{KL}(q(\omega|\theta) \parallel P(\omega|\mathcal{D})) = E[\log q - \log P] = \int q(\omega|\theta) \log\left(\frac{q(\omega|\theta)}{P(\omega|\mathcal{D})}\right) d\omega \tag{5}$$

Where the $E$ is the expectation operator and represented as difference between the natural logarithm of the variational posterior ($q$) and the actual posterior ($P$).

When the observed data is treated as constant (the inputs are treated as constant), the above equation is simplified into a function $L(\theta|\mathcal{D})$, that solely relies on the variables $\theta$ and $\mathcal{D}$ [14],

$$\begin{aligned} L(\theta|\mathcal{D}) &= D_{KL}(q(\omega|\theta) \parallel P(\omega)) - E_{q(\omega|\theta)}(\log P(\mathcal{D}|\omega)) \\ &= \int q(\omega|\theta)(\log q(\omega|\theta)) - \log P(\mathcal{D}|\omega) - \log P(\omega) \, d\omega \\ &= E_{q(\omega|\theta)}(\log q(\omega|\theta) - \log P(\mathcal{D}|\omega) - \log P(\omega)) \end{aligned} \tag{6}$$

In the equation (6), the initial term evaluates the difference between the variational and prior distributions, whereas the subsequent term represents the anticipated negative logarithmic probability (analogous to negative log likelihood loss term) based on the variational posterior. This function, Evidence Lower Bound (ELBO), $L(\theta|\mathcal{D})$, also known as the variational lower bound, can then be optimized with respect to $\theta$, the distribution of parameters (means and variances) to derive the best approximation of the true posterior.

We perform the optimization using a gradient-descent [9] and Resilient Back Propagation (RProp) [11] based "RMSprop" optimizer, allowing for the scalability of the variational Bayes mechanism to large and intricate models. The primary goal using the variational inference is to accurately approximate the true weights of the posterior distribution, treating the inputs as constant.

We provide the hyperparameters of the probabilistic ML algorithms in Table S2 and the model architecture summaries in Fig. S5 (a, b) of the Bayesian Neural Network approaches we used in this work.

**Table S3: Hyperparameters for probabilistic machine learning models**

| Probabilistic ML Algorithms | Hyperparameters |
| --- | --- |
| GPR | Matern (length_scale=1, nu=1.5) + WhiteKernel(noise_level=1), n_restarts_optimizer=50, random_state=2022 |
| BNN – trainable mean and variance | Number of hidden layers: 3, Number of neurons in the hidden layers: (24, 16, 8), Activation Function: Relu (only used in the hidden layers), Loss: "Negative Log Likelihood", Optimizer: "Adam" |
| BNN-working as an ensemble of networks | Number of hidden layers: 1, Number of neurons in the hidden layers: 8, Activation Function: Sigmoid, learning rate: 0.001, Loss: "Negative Log Likelihood", Optimizer: RMSProp. |

**(a) Model summary for the BNN – trainable mean and variance**

```
Layer (type)                    Output Shape              Param #
=================================================================
input_4 (InputLayer)            [(None, 16)]              0

dense_variational_1 (Dense      (None, 8)                 9452
Variational)

dense_1 (Dense)                 (None, 2)                 18

independent_normal_1 (Inde      ((None, 1),               0
pendentNormal)                   (None, 1))
=================================================================
Total params: 9470 (36.99 KB)
Trainable params: 9470 (36.99 KB)
Non-trainable params: 0 (0.00 Byte)
```

**(b) Model summary for the BNN – ensemble of networks**

```
Layer (type)                    Output Shape              Param #
=================================================================
dense_1 (Dense)                 (None, 24)                408

batch_normalization_3 (Batc     (None, 24)                96
hNormalization)

dense_2 (Dense)                 (None, 16)                400

batch_normalization_4 (Batc     (None, 16)                64
hNormalization)

dense_3 (Dense)                 (None, 8)                 136

batch_normalization_5 (Batc     (None, 8)                 32
hNormalization)

distribution_weights (Dense     (None, 2)                 18
)

output (MultivariateNormalT     ((None, 1),               0
riL)                             (None, 1))
=================================================================
Total params: 1,154
Trainable params: 1,058
Non-trainable params: 96
```

**Fig S3:** Model summaries for the probabilistic ML models we use for our work (a) Model summary for the preliminary approach - BNN with trainable mean and variance. (b) Model summary for our second approach where the BNN is working as the ensemble of networks.

## 4. Representation of uncertainties from Probabilistic Bayesian Neural Network with varying train-test split ratios

In Fig S4, we provide a plot illustrating the trends in aleatoric and epistemic uncertainty for the train-validation split ratios. We split the train dataset in different ratios ranging from 10% training split to 99% training split. It is noteworthy to emphasize that the 10% training data is exclusively utilized for observing the prediction distributions generated by the probabilistic network ensemble. It is widely recognized that training a model with a limited amount of data would inevitably result in an increase in model uncertainty, hence leading to an increase in epistemic uncertainty. Additionally, it is worth noting that this practice contributes to the amplification of noise within the dataset. This is mostly due to the uneven distribution of data between the training and testing sets, which subsequently leads to an escalation in aleatoric uncertainty. Based on the findings we present in Fig S4, we observe that the aleatoric uncertainty exhibits a decreasing pattern with intermittent increase while considering appropriate ratios of training fractions, such as 50% and 80% for training data. Additionally, we also observe a sudden increase in aleatoric uncertainty at 90% training percentage, when an uneven distribution of train-test data split is included. The observed pattern of increasing aleatoric uncertainty at 90% training data and then decreasing at 99% training data in the Probabilistic BNN might be due to few factors. At 90% training data, the model might have encountered more diverse and challenging data points, introducing additional variability in the predictions, thus leading to increased aleatoric uncertainty. This increase might signify that the model is struggling to capture complex patterns in this subset. As the training dataset further expands to 99%, the model likely gains a more comprehensive understanding of the data distribution, resulting in an improved predictive accuracy and reduced aleatoric uncertainty. Additionally, the increased training data may help the model better estimate the aleatoric uncertainty, leading to a more reliable representation. However, it's crucial here to consider the specifics of the dataset and model, as other factors like data quality, model architecture, or the nature of the training data might have also played a role in this observed behavior. Upon examining the outcomes pertaining to epistemic uncertainty, it becomes evident that there is a steady downward trajectory observed in conjunction with an increase in the number of training data points.

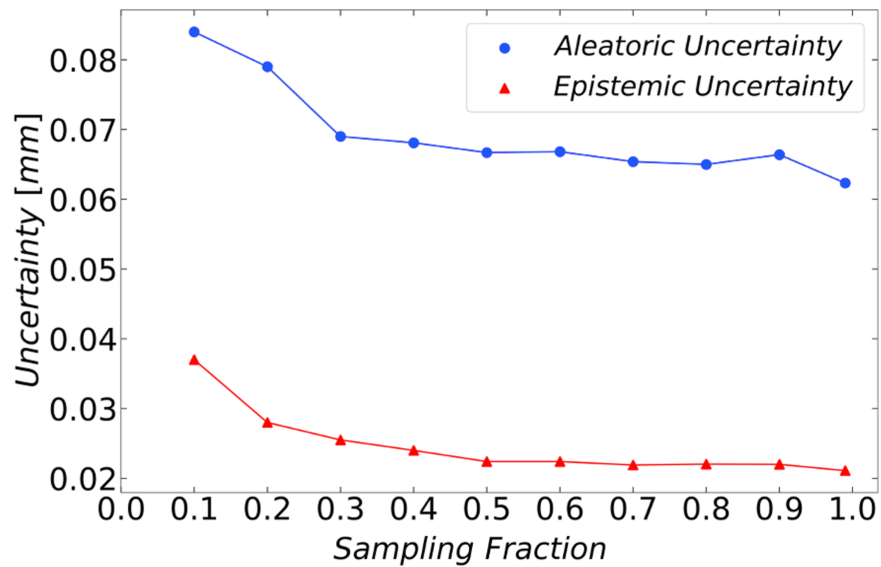

**Fig S4**: Representation of the aleatoric and epistemic uncertainties with varying train-validation split ratios. We observe that the aleatoric uncertainty exhibits a decreasing pattern with intermittent increase while considering appropriate ratios of training fractions, such as 50% and 80% for the training data. We observe a consistent downward trajectory in epistemic uncertainty in conjunction with an increase in the number of training data points.